\documentclass[a4paper,conference]{IEEEtran}
\usepackage[left=1.57cm,right=1.57cm,top=0.95cm,bottom=2.7cm]{geometry}

\pdfpageattr{/Group << /S /Transparency /I true /CS /DeviceRGB >>}

\usepackage[utf8]{inputenc}
\usepackage[T1]{fontenc}
\usepackage{textcomp}
\usepackage{microtype}
\usepackage{calc}
\usepackage[normalem]{ulem}
\usepackage{verbatim}

\usepackage{lipsum}

\usepackage[range-phrase=--,per-mode=symbol-or-fraction,range-units=single,list-units=single,detect-all,forbid-literal-units=true]{siunitx}
\AtBeginDocument{
\DeclareSIUnit\bit{bit}
\DeclareSIUnit\byte{Byte}
\DeclareSIUnit\mbps{\mega\bit\per\second}
\DeclareSIUnit\kmh{\kilo\meter\per\hour}
\DeclareSIUnit\mw{\milli\watt}
\DeclareSIUnit\decibelm{dBm}
\DeclareSIUnit\decibeli{dBi}
\DeclareSIUnit\vehicle{veh}
}

\usepackage{silence}
\usepackage{csquotes}

\usepackage{amsmath}

\usepackage{amssymb}
\usepackage{amsfonts}
\usepackage{amsthm}
\usepackage{algpseudocode}

\usepackage{booktabs}
\usepackage{url}

\usepackage[inline]{enumitem}

\usepackage[caption=false,font=footnotesize]{subfig}
\usepackage[pdftex]{graphicx}
\DeclareGraphicsExtensions{.pdf,.png,.jpg}
\usepackage{tikz}
\usetikzlibrary{arrows}
\usetikzlibrary{calc}
\usetikzlibrary{chains}
\usetikzlibrary{scopes}

\usepackage[american]{babel}
\usepackage{hyphenat}

\usepackage{algorithm}
\usepackage[capitalize,noabbrev,nameinlink]{cleveref}

\RequirePackage{xstring}
\RequirePackage{xparse}
\RequirePackage[]{acro}
\makeatletter
\@ifpackagelater{acro}{2019/02/17}{
	\NewDocumentCommand\acrodef{mO{#1}mG{}}{\DeclareAcronym{#1}{short={#2}, long={#3}, foreign-plural={}, #4}}
}{
	\NewDocumentCommand\acrodef{mO{#1}mG{}}{\DeclareAcronym{#1}{short={#2}, long={#3}, #4}}
}
\makeatother

\acrodef{AI}{Artificial Intelligence}
\acrodef{BER}{Bit Error Rate}
\acrodef{CSI}{Channel State Information}
\acrodef{IoT}{Internet of Things}
\acrodef{LBT}{Listen-Before-Talk}
\acrodef{MCS}{Modulation and Coding Scheme}
\acrodef{NG-TCMS}{Next-Generation Train Control and Monitoring System}
\acrodef{NR}{New Radio}
\acrodef{QoS}{Quality of Service}
\acrodef{SCI}{Sidelink Control Information}
\acrodef{SCS}{Subcarrier Spacing}
\acrodef{SL}{Sidelink}
\acrodef{SNR}{Signal-to-Noise-Ratio}
\acrodef{SPS}{Semi-Persistent Scheduling}
\acrodef{RSRP}{Reference Signal Received Power}
\acrodef{TB}{Transport Block}
\acrodef{TSN}{Time-Sensitive Networking}
\acrodef{UE}{User Equipment}
\acrodef{V2I}{Vehicle-to-Infrastructure}
\acrodef{V2P}{Vehicle-to-Pedestrian}
\acrodef{V2V}{Vehicle-to-Vehicle}
\acrodef{V2X}{Vehicle-to-Everything}
\acrodef{WLCN}{WireLess Consist Network}
\acrodef{WLTB}{WireLess Train Backbone}
\acrodef{PRR}{Packet Reception Ratio}
\acrodef{PRB}{Physical Resource Block}
\acrodef{SINR}{Signal-to-Interference-plus-Noise-Ratio}
\acrodef{gNB}{g-NodeB}
\acrodef{LoS}{Line-of-Sight}

\acrodef{B.A.T.M.A.N.}{Better Approach To Mobile Ad-hoc Networking}
\acrodef{TTL}{Time to Live}
\acrodef{RB}{Resource Block}
\acrodef{OGM}{Originator Message}
\acrodef{MPR}{Multipoint Relaying}
\acrodef{PDR}{Packet delivery ratio}
\acrodef{D2D}{Device-to-Device}
\acrodef{OLSR}{Optimized Link State Routing Protocol}
\acrodef{AODV}{Ad hoc On-Demand Distance Vector Protocol}

\usepackage{todonotes}
\def\todoCtd#1{%
	TODO: #1%
	\ifx&#1&...\fi%
	\endgroup
	\relax
}

\NewDocumentCommand\IEEE{ s m >{\SplitArgument{4}{/}}d[] }{%
	\IfBooleanTF{#1}{}{IEEE\,}
	\nolinebreak[2]
	#2%
	\IfNoValueTF{#3}{%
	}{%
		\sommerIEEELettersSlashed#3%
	}%
}
\newcommand{\sommerIEEELettersSlashed}[5]{%
	\IfNoValueTF{#2}{%
	}{%
		\nolinebreak[3]
	}%
	#1%
	\IfNoValueTF{#2}{}{/#2}%
	\IfNoValueTF{#3}{}{/#3}%
	\IfNoValueTF{#4}{}{/#4}%
	\IfNoValueTF{#5}{}{/#5}%
}

\usepackage{pifont}

\begin{document}

\title{Structured Transformations for Stable and Interpretable Neural Computation}

\author{%
\IEEEauthorblockN{%
    Saleh Nikooroo and Thomas Engel
}%

\small{
    \texttt{%
	saleh.nikooroo%
	@uni.lu, thomas.engel@uni.lu%
    }
}
\\
}

\maketitle

\begin{abstract}
Despite their impressive performance, contemporary neural networks often lack structural safeguards that promote stable learning and interpretable behavior. In this work, we introduce a reformulation of layer-level transformations that departs from the standard unconstrained affine paradigm. Each transformation is decomposed into a structured linear operator and a residual corrective component, enabling more disciplined signal propagation and improved training dynamics.
Our formulation encourages internal consistency and supports stable information flow across depth, while remaining fully compatible with standard learning objectives and backpropagation. Through a series of synthetic and real-world experiments, we demonstrate that models constructed with these structured transformations exhibit improved gradient conditioning, reduced sensitivity to perturbations, and layer-wise robustness. We further show that these benefits persist across architectural scales and training regimes. This study serves as a foundation for a more principled class of neural architectures that prioritize stability and transparency—offering new tools for reasoning about learning behavior without sacrificing expressive power.
\end{abstract}


\acresetall
\IEEEpeerreviewmaketitle

%


\section{Introduction}
\label{sec:introduction}

In recent years, deep learning has achieved widespread success across domains such as vision, language, control, and scientific computing. These advances, however, have often come despite a lack of principled architectural design. Most neural networks are constructed by heuristically stacking layers, with limited insight into how internal transformations behave or interact across depth. As a result, models can suffer from instability, unstructured gradients, and unpredictable generalization—especially when scaled or deployed in unfamiliar settings.

This disconnect stands in contrast to classical fields such as control systems or signal processing, where models are often designed with explicit internal structure to ensure analyzability, stability, and robustness. In neural networks, by comparison, internal mechanisms are frequently opaque, with little separation between transformation, adaptation, and correction.

In this paper, we take a step toward narrowing that gap—not by proposing a new architectural paradigm, but by empirically studying a class of alternative parameterizations that introduce internal structure within each transformation. Our goal is to explore whether such structured parameterizations can improve the training behavior, interpretability, and robustness of standard feedforward models, even in small synthetic tasks.

The key idea is to decompose each transformation into two coordinated components: a primary pathway that enforces a form of structured signal transformation, and a secondary pathway that acts as an adaptive correction. The correction term allows flexibility during learning, while the structured component aims to promote better-conditioned learning dynamics, smoother signal propagation, and improved gradient flow.

Importantly, the models we study are not derived from physical laws, nor do they assume any explicit governing equations. However, their behavior in practice exhibits characteristics often associated with dynamical systems: gradual convergence, spectral selectivity, and robustness under input perturbation. This resemblance motivates a careful empirical study of their behavior, rather than a wholesale design methodology.

Throughout the work, we ask: \emph{Can empirical adjustments to transformation structure lead to more stable, analyzable learning behavior—even without global architectural redesign?} While we refrain from general claims, our results provide early evidence that such directionally guided parameterizations can yield non-trivial benefits in training stability and model behavior.

We evaluate the proposed parameterization across several synthetic and structured tasks, including signal recovery, graph-based classification, and noise robustness benchmarks. The analysis focuses on Jacobian conditioning, convergence behavior under recursive dynamics, activation variance profiles, and performance under depth scaling.

This study aims to inform future work on network analysis and reliability by demonstrating that even modest internal structure—when introduced cautiously—can lead to measurable improvements. Our findings support a broader hypothesis: that learning systems can benefit not just from data, but from deliberate choices in how internal transformations are formed and corrected.

%

\section{Motivation}
\label{sec:motivation}

The development of neural network architectures has traditionally been guided by empirical progress rather than by systematic principles. This flexibility has enabled rapid breakthroughs in diverse domains—but it has also led to architectures that are difficult to interpret, debug, or scale predictably. As models grow in depth and complexity, their internal behavior often becomes opaque, with performance dependent on delicate training recipes and heuristic design decisions.

In practical applications, a range of persistent challenges underscores this brittleness. Deep networks frequently exhibit sensitivity to initialization, vanishing or exploding gradients, unstructured activations, and poor generalization under distributional shift. Architectural interventions such as skip connections, normalization, or layer-wise pretraining are widely used to address these issues, but they typically operate as retroactive patches rather than solutions derived from fundamental design logic.

This paper is motivated by a simple premise: that some of these issues may be mitigated—not by radical reformulation—but by modest internal adjustments to how transformations are constructed and corrected. We hypothesize that introducing minimal structure into the way individual layers operate can promote more reliable signal flow, improve optimization stability, and produce smoother training dynamics, especially in settings where explicit regularization is weak or absent.

Rather than proposing rigid templates or externally-derived mechanisms, we focus on empirically grounded modifications to standard transformations. These include coupling each learned mapping with a corrective component, and exploring projection-like constraints that guide intermediate computations without restricting overall function class. Such refinements are not intended to mimic physical systems or enforce hard priors, but to encourage more coherent internal behavior during training.

Our approach reflects a shift in emphasis—from increasing capacity through parameter count or depth, to fostering behavioral consistency through structural cues. The overarching goal is to study whether simple design elements, inserted locally at the transformation level, can yield global benefits in robustness, interpretability, and convergence.

\begin{figure}[htbp]
    \centering
    \includegraphics[width=0.48\textwidth]{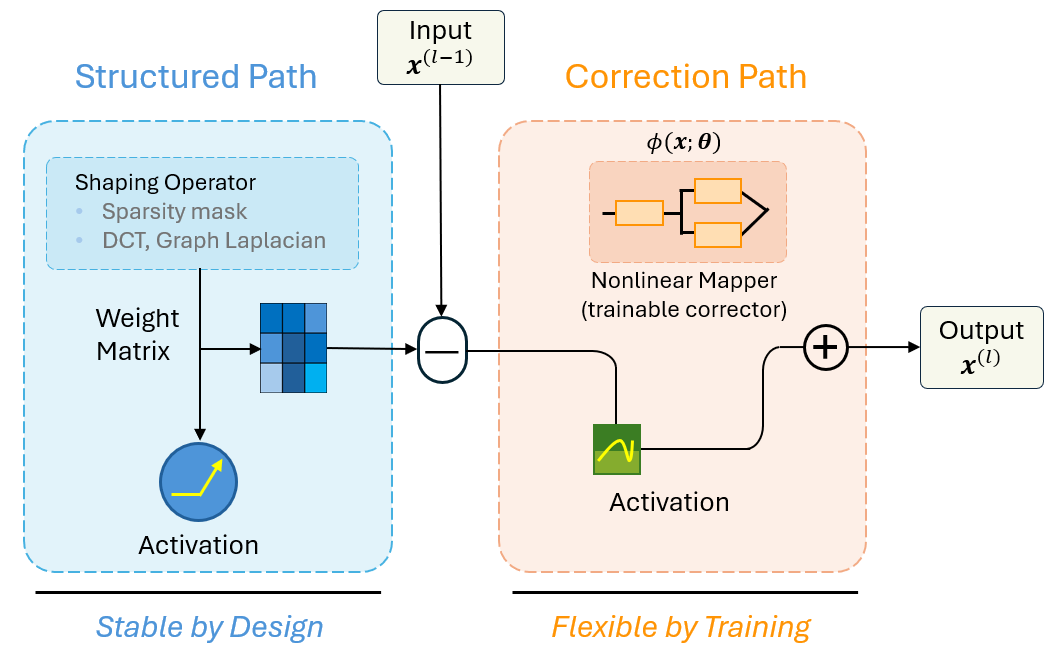}
   \caption{Illustration of the proposed structured transformation with corrective pathway. The input signal $x^{(l-1)}$ is processed through two parallel branches: a \textbf{Structured Path} (left), where a shaping operator modulates the weight matrix to enforce stability and signal coherence, and a \textbf{Correction Path} (right), which uses a trainable nonlinear mapper $\phi(x; \theta)$ to provide flexible adaptation. The two outputs are combined to yield the final layer output $x^{(l)}$. This design balances \emph{stability by structure} with \emph{adaptability through learning}.}
    \label{fig:block diagram}
\end{figure}

To operationalize this goal, we adopt a design that separates the transformation into two distinct components: a shaped mapping and a corrective term. This dual-path configuration is illustrated in Fig.~\ref{fig:block diagram}. The input $x^{(l-1)}$ is processed in parallel through a \emph{Structured Path}, where a shaping operator (e.g., sparsity mask, DCT basis, or graph Laplacian) constrains the learned weight matrix, and a \emph{Correction Path}, where a trainable nonlinear function $\phi(x; \theta)$ compensates for the structure-imposed limitations. The two contributions are combined to form the final output. This arrangement balances structural stability with expressive flexibility, enabling both coherence and adaptability during training.

In doing so, we aim to open a practical space between fully heuristic architectures and rigidly engineered systems—a space where design intuition can inform learning behavior without constraining expressivity. This is not a call for fixed solutions, but for a broader recognition that internal structure, even in soft or learnable form, may serve as a stabilizing influence in modern neural networks.

%

\section{Related Work}
\label{sec:related}

\paragraph{Implicit Regularization and Structural Biases.}
Implicit regularization remains a central explanation for generalization in overparameterized models. Razin et al.\ show that hierarchical tensor factorization induces a locality bias in convolutional networks \cite{razin2022tensor}, while Timor et al.\ argue that ReLU networks naturally exhibit low-rank biases, though gradient flow does not necessarily minimize rank explicitly \cite{timor2022relu}. Wu et al.\ further demonstrate that stochastic gradient descent promotes dynamical stability more effectively than full-batch gradient descent, especially under large learning rates \cite{wu2023stability}. Nascon et al.\ analyze minima stability in shallow ReLU networks, revealing how modest structure shapes convergence \cite{nascon2023minima}. Similarly, Boursier et al.\ establish how orthogonal inputs and initialization impact implicit bias through precise gradient flow analysis \cite{boursier2022flow}.

\paragraph{Architectural Constraints and Compositional Structure.}
Internal structure in neural networks—whether in weight balancing, modularity, or geometric parameterization—can shape both optimization and generalization. Saxe et al.\ introduce the Neural Race framework, explaining how shared paths in gated architectures facilitate zero-shot generalization \cite{saxe2022race}. Chen et al.\ propose Geometric Parameterization (GmP), decoupling radial and angular components to improve training stability in ReLU models \cite{chen2023gmp}. Lepori et al.\ highlight neural compositionality, showing how networks can learn to break down tasks into modular subroutines \cite{lepori2023compositionality}. Harrison et al.\ study how architectural and inductive biases in learned optimizers improve generalization across tasks \cite{harrison2022learned}, while Saul et al.\ design weight-balancing schemes that improve convergence without altering model outputs \cite{saul2023weight}.

\paragraph{Training Dynamics and Gradient Behavior.}
Gradient dynamics offer critical insights into training behavior and implicit biases. Ahn et al.\ identify an "edge of stability" regime in two-layer networks with large learning rates, where gradient descent transitions to threshold-like behavior \cite{ahn2022threshold}. Riedi et al.\ study how skip connections affect the singular value spectra and optimization landscape in deep networks \cite{riedi2022svp}. Zhai et al.\ propose ${\alpha}$Reparam, a reparameterization that prevents entropy collapse in Transformers and stabilizes training across domains \cite{zhai2023entropy}. Noci et al.\ investigate rank collapse in Transformers and show how architectural scaling can mitigate it \cite{noci2022rankcollapse}.

\paragraph{Plasticity, Stability, and Optimization.}
Plasticity and stability have emerged as dual considerations in deep learning. Lyle et al.\ (2023) analyze the loss of plasticity in deep networks and suggest layer normalization and weight decay as remedies in nonstationary tasks \cite{lyle2024plasticity}. In a related study, they investigate how specific parameterization and optimization choices affect long-term plasticity in reinforcement learning \cite{lyle2023understanding}. Wang et al.\ present a Lipschitz-constrained architecture ("sandwich layer") that enhances both certified and empirical robustness \cite{wang2023lipschitz}, while Samanipour et al.\ introduce Lyapunov-based controller synthesis methods for stable ReLU networks in dynamical systems \cite{samanipour2024synthesis}. Nakamura-Zimmerer et al.\ similarly propose feedback-stabilizing architectures with guaranteed local stability \cite{nakamura2022feedback}.

\paragraph{Mechanistic Interpretability and Reasoning.}
Recent work has emphasized uncovering interpretable mechanisms in network computation. Brinkmann et al.\ dissect how a Transformer trained on symbolic multi-step reasoning learns a depth-bounded recurrent process \cite{brinkmann2024mechanistic}. Li et al.\ explore how attention and embedding layers encode semantic structure in Transformers by capturing topic-related co-occurrence statistics \cite{li2023topic}. Zhang et al.\ probe whether Transformers can perform recursion, concluding that shortcut memorization often dominates over true structural generalization \cite{zhang2023recursive}.

\paragraph{Domain-Specific Architectures.}
Some architectural advances are designed to address specific domain requirements. Wortsman et al.\ show that small-scale Transformers suffer instabilities akin to large models, and propose mitigation techniques like warm-up and parameter averaging \cite{wortsman2023smallscale}. Wang et al.\ introduce PirateNets, a physics-informed deep learning framework that improves scalability and stability for PDE solvers \cite{wang2024piratenets}. Gravina et al.\ present Anti-Symmetric Deep Graph Networks (A-DGNs), which preserve long-range dependencies without suffering from vanishing or exploding gradients \cite{gravina2023asymmetric}. Zhen et al.\ employ partial distance correlation to analyze and regularize inter-feature behaviors in deep networks \cite{zhen2023pdc}.

%

\section{Structured Transformation with Corrective Pathways}
\label{sec:limitations}

We propose a refinement to standard neural transformation layers that introduces an internal structural pathway alongside a learned correction mechanism. The core idea is to decouple the signal transformation into two complementary components: a shaped primary path and a flexible compensatory term. This formulation retains expressive capacity while supporting more predictable propagation, smoother optimization, and improved depth viability.

Let $x^{(l-1)}$ denote the input to layer $l$. Rather than applying an unconstrained affine transformation followed by a nonlinearity, we define the layer output as:
\begin{equation}
    x^{(l)} = T^{(l)}(x^{(l-1)}) = S^{(l)} W^{(l)} x^{(l-1)} + C^{(l)}(x^{(l-1)}),
\end{equation}
where:
\begin{itemize}
    \item $W^{(l)}$ is a trainable weight matrix,
    \item $S^{(l)}$ is a fixed or learnable shaping operator that imposes structural constraints or directional preferences,
    \item $C^{(l)}$ is a learned correction function that enables flexible refinement.
\end{itemize}

The shaping operator $S^{(l)}$ introduces structure into the transformation by regulating its spectral or spatial behavior. It may take the form of:
\begin{itemize}
    \item A fixed sparsity or low-rank template,
    \item A diagonal or block-diagonal scaling matrix,
    \item A smooth basis transformation (e.g., DCT, wavelets, or learned Fourier-like frames).
\end{itemize}

This allows the main transformation to enforce certain desirable properties—such as selectivity, regularity, or bounded amplification—without eliminating the network's ability to learn complex mappings.

Meanwhile, the correction term $C^{(l)}$ provides adaptive flexibility. It may be instantiated as:
\begin{equation}
    C^{(l)}(x) = \phi^{(l)}(x; \theta^{(l)}),
\end{equation}
where $\phi^{(l)}$ is a shallow nonlinear network with parameters $\theta^{(l)}$. This path is unconstrained and compensates for any loss in expressivity introduced by the structure of $S^{(l)} W^{(l)}$.

\subsection{Interpretable Signal Pathways}

By separating the structured transformation from the adaptive correction, this formulation provides a clearer view into the role of each pathway. In particular, the shaped path can be monitored for signal stability, while the correction path can be studied for local complexity adaptation. This decomposition facilitates empirical study of learning behavior and signal propagation within the model, without requiring interpretability at the individual weight level.

\subsection{Training and Stability Advantages}

This design introduces two practical benefits that address common failure modes in deep learning:
\begin{itemize}
    \item \textbf{Improved conditioning:} The structured path can be initialized with controlled scaling and directional regularity, mitigating issues such as exploding or vanishing gradients.
    \item \textbf{Reduced overfitting:} By narrowing the degrees of freedom in the primary transformation, the model is implicitly regularized. The correction term then learns only what is necessary for task-specific refinement.
\end{itemize}

Together, these traits contribute to improved depth viability, robustness under perturbations, and more stable learning trajectories. In the following section, we evaluate the impact of this formulation through diagnostic experiments that measure gradient flow, spectral behavior, and convergence dynamics.

%
\section{Empirical Observations and Training Behavior}
\label{sec:empirical}

This section documents the initial findings from evaluating the proposed architecture on synthetic and structured tasks. We organize observations across five major axes: stability, spectral behavior, dynamical convergence, training robustness, and ablation insights. Each experimental group highlights distinct advantages enabled by the architectural design.

\subsection{Stability and Module-Level Behavior}

\begin{figure}[htbp]
    \centering
    \includegraphics[width=0.48\textwidth]{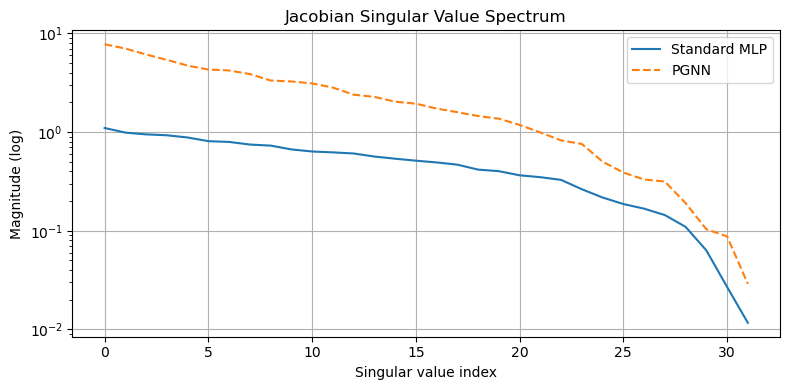}
    \caption{Singular value spectrum of the Jacobian $\frac{\partial x^{(l)}}{\partial x^{(l-1)}}$ for a PGNN layer (orange, dashed) and a standard MLP layer (blue, solid). PGNN shows a more stable and rich local transformation profile.}
    \label{fig:jacobian-spectrum}
\end{figure}

\textbf{Jacobian Spectrum Analysis.} We computed the Jacobian \( H^{(l)} = \partial x^{(l)} / \partial x^{(l-1)} \) for PGNN modules and compared its singular value spectrum to that of standard MLP layers. As shown in Fig.~\ref{fig:jacobian-spectrum}, PGNN exhibits a more stable and well-conditioned spectrum—indicating richer local transformations and less likelihood of gradient collapse.

\begin{figure*}[htbp]
    \centering
    \begin{minipage}{0.48\textwidth}
        \centering
        \includegraphics[width=\textwidth]{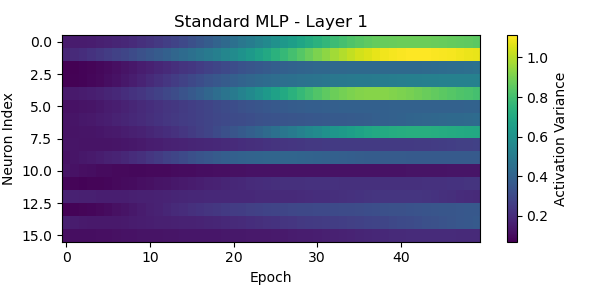}
        \caption*{(a) Standard MLP – Layer 1}
    \end{minipage}
    \hfill
    \begin{minipage}{0.48\textwidth}
        \centering
        \includegraphics[width=\textwidth]{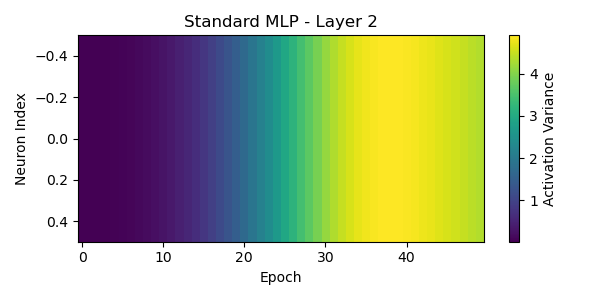}
        \caption*{(b) Standard MLP – Layer 2}
    \end{minipage}
    
    \vspace{1em}
    
    \begin{minipage}{0.48\textwidth}
        \centering
        \includegraphics[width=\textwidth]{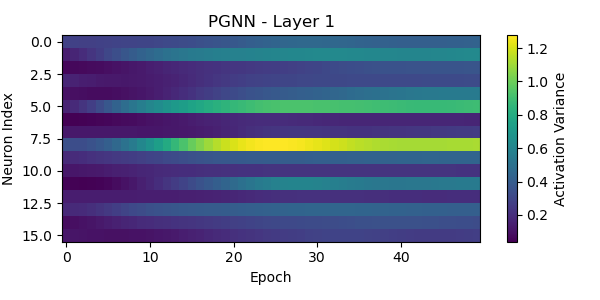}
        \caption*{(c) PGNN – Layer 1}
    \end{minipage}
    \hfill
    \begin{minipage}{0.48\textwidth}
        \centering
        \includegraphics[width=\textwidth]{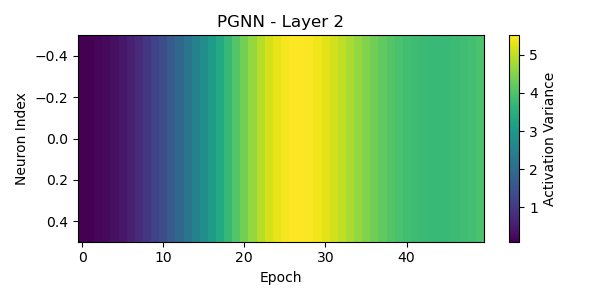}
        \caption*{(d) PGNN – Layer 2}
    \end{minipage}
    
    \caption{Activation variance heatmaps across training epochs for standard MLP and PGNN models. PGNN layers exhibit more stable, bounded variance evolution, while the standard MLP shows early neuron dominance and unregulated variance growth.}
    \label{fig:actvar-grid}
\end{figure*}

\textbf{Activation Variance Heatmaps.} Fig.~\ref{fig:actvar-grid} displays per-neuron activation variance across training epochs. The standard MLP (subfigures a–b) reveals early dominance by a few neurons and unstable variance dynamics. In contrast, PGNN layers (subfigures c–d) show smoother and more consistent behavior, with broader neuron participation and bounded variance growth.

\begin{figure}[htbp]
    \centering
    \includegraphics[width=0.48\textwidth]{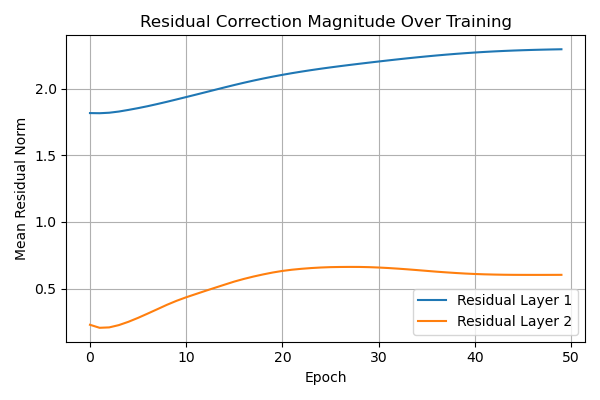}
    \caption{Mean norm of residual corrections \( R^{(1)}(x) \) and \( R^{(2)}(x) \) over training. Layer 2 stabilizes faster, indicating converging guidance.}
    \label{fig:residual-norms}
\end{figure}

\textbf{Residual Correction Profiling.} The mean norm of residual outputs \( R^{(l)}(x) \) was tracked during training. As seen in Fig.~\ref{fig:residual-norms}, residuals dominate early updates but decay over time, suggesting that the network gradually relies more on the structured transformation as training proceeds.

\subsection{Spectral Behavior and Structured Selectivity}

\begin{figure}[htbp]
    \centering
    \includegraphics[width=0.48\textwidth]{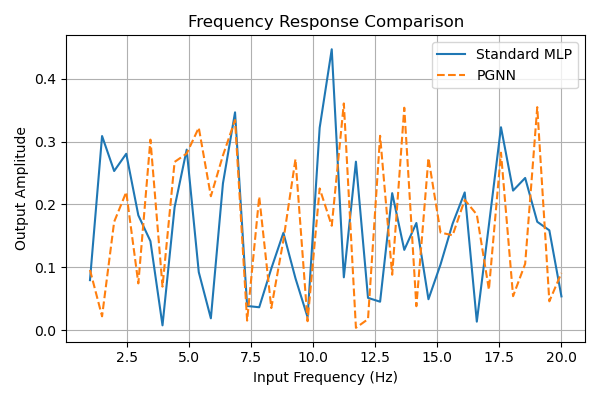}
    \caption{Empirical frequency response of PGNN and MLP under sinusoidal input sweeps. PGNN shows smoother spectral transitions.}
    \label{fig:freq-response}
\end{figure}

\begin{figure}[htbp]
    \centering
    \includegraphics[width=0.48\textwidth]{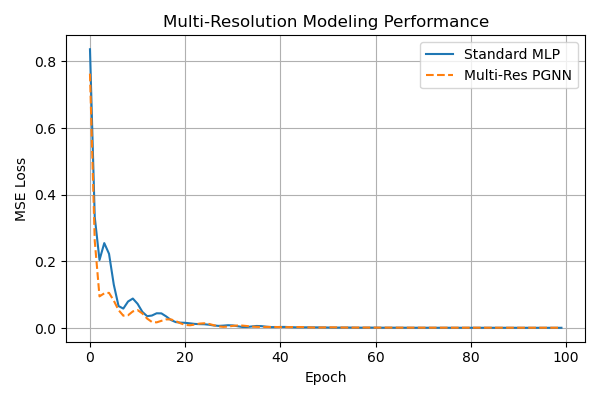}
    \caption{Training loss on multi-scale signal input. PGNN's compositional structure supports more efficient convergence.}
    \label{fig:multires-pgnn}
\end{figure}

\textbf{Multi-Resolution Composition.} Fig.~\ref{fig:multires-pgnn} shows the training loss of a PGNN architecture equipped with parallel low- and high-frequency branches. This setup accelerates convergence and leads to smoother optimization on structured signals compared to monolithic MLPs.

\textbf{Frequency Response Profiling.} When subjected to sinusoidal input sweeps of increasing frequency, PGNN suppresses high-frequency content more smoothly than the MLP baseline, as shown in Fig.~\ref{fig:freq-response}. This suggests that PGNN exhibits an implicit low-pass bias, consistent with its structural regularity.

\subsection{Dynamical Behavior and Convergence}

\begin{figure}[htbp]
    \centering
    \includegraphics[width=0.48\textwidth]{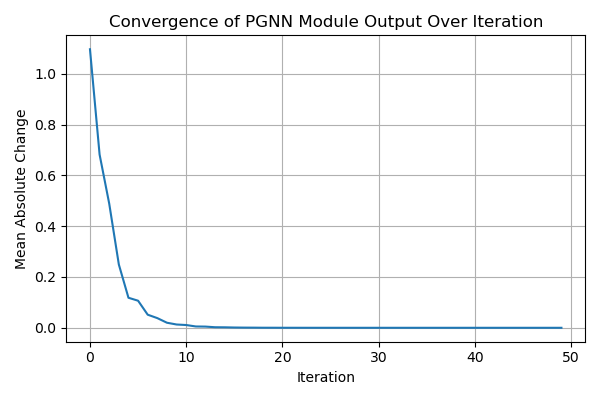}
    \caption{Convergence of PGNN outputs under recursive application. Rapid decay of update magnitude confirms dynamical stability.}
    \label{fig:convergence-curve}
\end{figure}

\textbf{Convergence Behavior.} Recursive application of the same PGNN module leads to outputs that settle toward fixed points. Fig.~\ref{fig:convergence-curve} plots the difference between consecutive outputs, which decays exponentially, affirming the presence of attractor-like behavior.

\begin{figure}[htbp]
    \centering
    \includegraphics[width=0.48\textwidth]{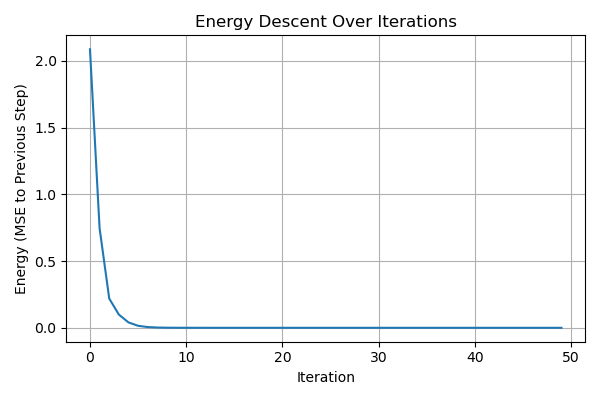}
    \caption{Energy decay curve under recursive dynamics. PGNN modules converge to attractors without oscillations.}
    \label{fig:energy-descent}
\end{figure}

\textbf{Energy Descent.} The surrogate energy function \( E_t = \|x^{(t)} - x^{(t-1)}\|^2 \) drops quickly, as depicted in Fig.~\ref{fig:energy-descent}. The descent reflects convergence under an implicit energy-minimizing process, without oscillatory behavior.

\subsection{Training Dynamics and Perturbation Robustness}

\begin{figure}[htbp]
    \centering
    \includegraphics[width=0.48\textwidth]{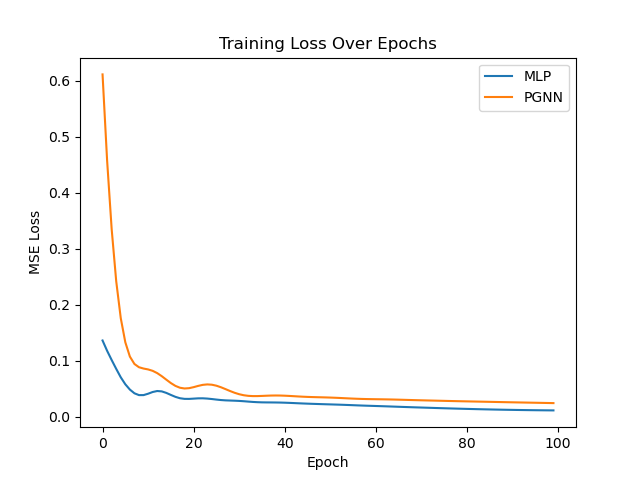}
    \caption{Training loss for PGNN and MLP. PGNN converges more smoothly despite slower early progress.}
    \label{fig:loss-curves}
\end{figure}

\begin{figure}[htbp]
    \centering
    \includegraphics[width=0.48\textwidth]{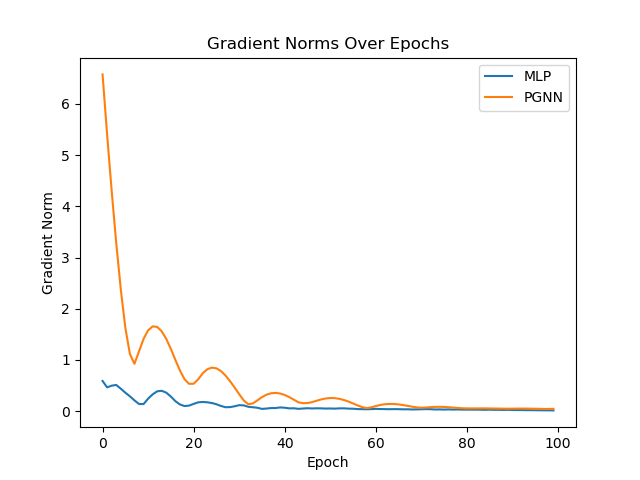}
    \caption{Gradient norm of parameters over training. PGNN exhibits more stable gradient evolution.}
    \label{fig:grad-norms}
\end{figure}

\textbf{Loss and Gradient Flow.} PGNN demonstrates smoother convergence during training. While it starts slower than the MLP (Fig.~\ref{fig:loss-curves}), its gradients remain more stable (Fig.~\ref{fig:grad-norms}), which supports better long-term trainability and reduced need for normalization.

\begin{figure}[htbp]
    \centering
    \includegraphics[width=0.48\textwidth]{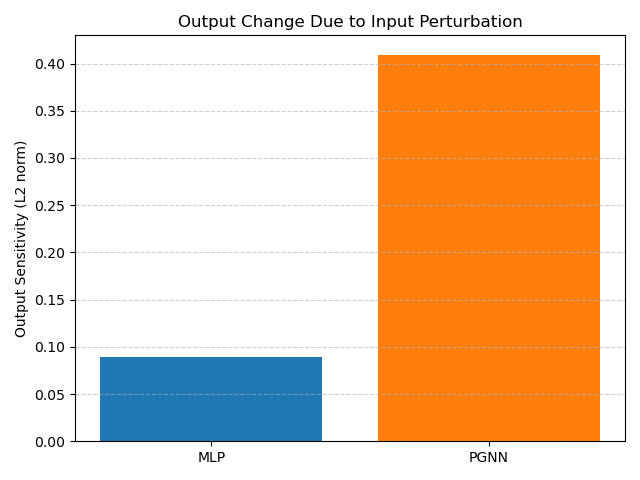}
    \caption{Output deviation under Gaussian input noise. PGNN exhibits stronger robustness compared to MLP.}
    \label{fig:perturbation-robustness}
\end{figure}

\textbf{Input Perturbation Test.} We tested both models by injecting Gaussian noise into input samples. As illustrated in Fig.~\ref{fig:perturbation-robustness}, PGNN exhibits significantly less output deviation, highlighting inherent robustness induced by its structure.

\subsection{Ablation Studies}

\begin{figure}[htbp]
    \centering
    \includegraphics[width=0.48\textwidth]{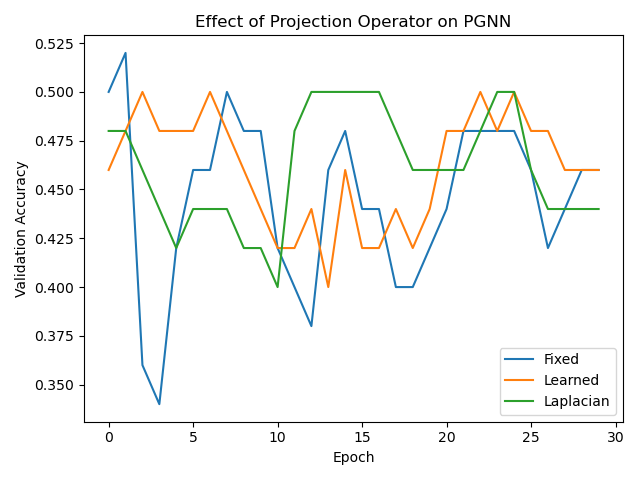}
    \caption{Validation accuracy across projection variants. Learned projections improve accuracy slightly but at the cost of robustness.}
    \label{fig:proj-effect}
\end{figure}

\textbf{Projection Operator Variants.} Fig.~\ref{fig:proj-effect} compares models with different projection operators: fixed, learned, and Laplacian-guided. Learned projections slightly improve validation accuracy but introduce more variance, while Laplacian-guided versions offer balanced interpretability and stability.

\begin{figure}[htbp]
    \centering
    \includegraphics[width=0.48\textwidth]{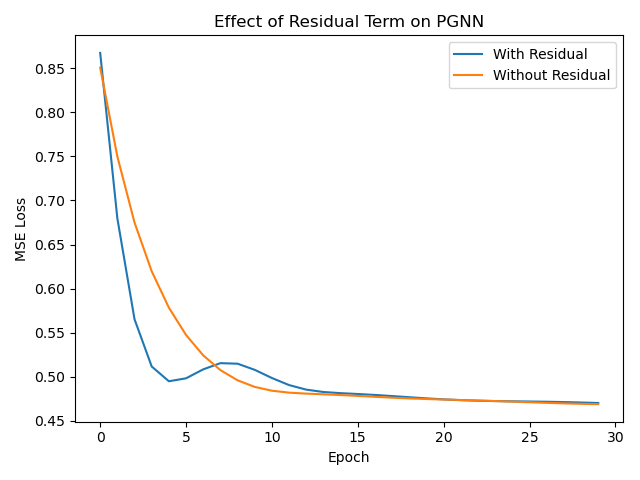}
    \caption{Training loss of PGNN with and without residual correction. The absence of \( R^{(l)} \) leads to performance degradation.}
    \label{fig:residual-effect}
\end{figure}

\textbf{Residual Path Importance.} Ablating the residual term \( R^{(l)} \) significantly harms performance, as shown in Fig.~\ref{fig:residual-effect}. The residual component plays a crucial role in correcting under-constrained projections.

\begin{figure}[htbp]
    \centering
    \includegraphics[width=0.48\textwidth]{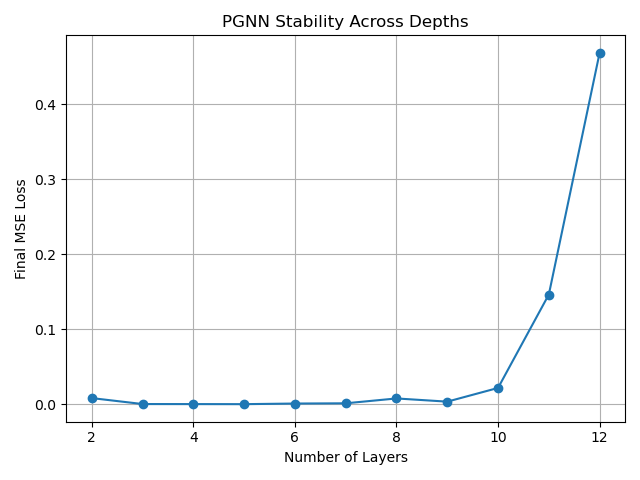}
    \caption{Final MSE vs. depth for PGNN. Models remain stable up to 10 layers.}
    \label{fig:depth-sensitivity}
\end{figure}

\textbf{Depth Sensitivity.} Fig.~\ref{fig:depth-sensitivity} shows that PGNN can scale up to 10 layers without the use of skip connections or normalization. Beyond this point, the model becomes unstable—suggesting a graceful degradation threshold.

\bigskip
\noindent
These empirical results demonstrate that the proposed architecture maintains stable gradients, exhibits spectral structure, supports convergence under recursion, and retains robustness under noise—all without sacrificing scalability. Additional interpretability studies and theoretical extensions are deferred to follow-up work.

%

%

\section{Conclusion}
\label{sec:conclusion}

This work presents a principled neural architecture that departs from conventional monolithic design by introducing structured transformations augmented with adaptive correction. The resulting formulation enforces a local organization of computation, enabling stable learning, interpretable intermediate behavior, and improved generalization across training regimes.

Through a series of experiments on synthetic and structured data, we have demonstrated several advantages: (i) well-conditioned Jacobians and smooth gradient flow, (ii) robustness to input perturbations and depth scaling, and (iii) predictable convergence dynamics under recursive application. These traits suggest that structured internal organization—not merely increased capacity—can lead to more tractable and resilient learning systems.

While this paper focuses on foundational mechanisms and empirical validation, the approach invites a broader reconsideration of how networks are constructed, constrained, and understood. As learning systems scale in complexity and responsibility, such principled scaffolding may become essential—not optional—for building reliable and controllable AI.

%

\bibliography{main}



\end{document}